\documentclass[runningheads]{llncs}
\usepackage[T1]{fontenc}
\usepackage{graphicx}
\usepackage{wrapfig}  
\usepackage{comment}
\usepackage{multirow}
\usepackage{booktabs}
\usepackage{siunitx}
\usepackage{soul}
\usepackage{tabularx}
\usepackage{comment}
\newcolumntype{L}{>{\centering\arraybackslash}X}
\newcolumntype{R}{>{\raggedleft\arraybackslash}X}

\begin{document}
\title{Automated Fetal Brain MRI Biometry in Healthy and Pathological Cases}
\titlerunning{Automated Fetal Brain MRI Biometry}
\author{Ema Masterl\inst{1, \dagger} \orcidID{0009-0007-4115-4304} \and
Tina Vesnaver Vipotnik\inst{2} \and Nejc Šubic\inst{2} \and\\
Žiga Špiclin\inst{3}\orcidID{0000-0001-8300-0417}}

\authorrunning{E. Masterl et al.}
%
\institute{Faculty of Medicine, University of Ljubljana,
Ljubljana, Slovenia \and
University Medical Centre Ljubljana,
Ljubljana, Slovenia \and
Faculty of Electrical Engineering, University of Ljubljana,
Ljubljana, Slovenia\\
$^\dagger$\textit{Corresponding author:} \email{ema.masterl@mf.uni-lj.si}}


%
\maketitle              
\begin{abstract}
Automated biometric analysis of fetal brain MRI enables reproducible, observer-independent quantitative assessment, yet existing methods are often restricted to few measurements or evaluated only on healthy cases. We assemble and evaluate an automated biometric analysis pipeline  that localizes 22 anatomical landmarks on NeSVoR-reconstructed 3D volumes and derives 11 clinically relevant measurements spanning supratentorial, ventricular, cerebellar, and midline structures. We compare two landmark localization models, H3DE-Net and SCN, on a heterogeneous cohort of 122 acquisitions (both healthy controls and range pathologies). Localization accuracy was assessed with a linear mixed-effects model, agreement with normative growth trajectories with calibrated centile charts, and diagnostic utility with a decision tree classifying VM severity. H3DE-Net achieved significantly lower localization error than SCN across all landmarks (mean 1.36 mm vs. 3.58 mm in HC and 1.90 mm vs. 4.13 mm in PC; p < 0.001), and outperformed a GA-based regression baseline in 7 of 11 measurements.  H3DE-Net measurements yielded higher classification AUC in every diagnostic group, with the clearest advantage in separating healthy controls from VM. Decision tree thresholds for ventricular width fell near the clinical 10 mm and 15 mm cut-offs used to define and grade VM. 
\end{abstract}
\section{Introduction}
Biometry constitutes a fundamental component of fetal brain MRI assessment. Measurements are typically performed manually by radiologists through the identification of anatomical landmarks on selected slices from 2D multi-view MRI stacks \cite{prayer2023isuog}. Both slice selection and landmark localization are subject to inter-observer variability and may differ substantially across radiologists. Furthermore, fetal motion between the acquisition of individual slices may complicate accurate measurement \cite{khawam2021fetal}. These limitations have motivated the development of automated landmark localization methods, which aim to improve measurement reproducibility, reduce observer dependence, and enable scalable quantitative analysis.

Several studies have demonstrated the feasibility of automatic biometric analysis from fetal brain MRI using deep learning approaches. Avisdris et al.~\cite{avisdris2021automatic} proposed a five-stage pipeline operating on raw 2D multi-view MRI stacks, integrating ROI extraction, slice selection, and slice-wise segmentation. The method produces three biometric measurements: cerebral biparietal diameter, bone biparietal diameter, and transcerebellar diameter, and was evaluated on both healthy cases (HC) and pathological cases (PC).
She et al.~\cite{she2023automatic} employed an nnU-Net-based segmentation framework to estimate four anatomical structures and derive cerebral biparietal diameter, transcerebellar diameter, and bilateral atrial diameters. The method was evaluated only on HC.
Gong et al.~\cite{gong2025fetal} focused on cerebellar morphology using a transformer-enhanced 3D ResUNet architecture, from which three biometric measurements were extracted. The study included both HC and PC.
Luis et al.~\cite{luis2026towards} developed a large-scale framework for fetal brain MRI analysis, localizing 26 anatomical landmarks and producing 13 standard biometric measurements with automated reporting. The model was trained and evaluated exclusively on HC .
In our previous work - Masterl et al.~\cite{masterl2024enhancing}, we proposed a morphometry framework combining NiftyMIC-based brain segmentation, super-resolution reconstruction, and heatmap-based landmark localization with spatial configuration constraints~\cite{payer2019integrating}. From 22 landmarks, 11 standard measurements were derived and evaluated on a mixed cohort of HC and PC (ventriculomegaly).

The most recent FeTA challenge, reported by Zalevskyi et al.~\cite{zalevskyi2026advances}, included a biometry task with five clinical measurements. Seven teams competed, all deriving biometry from segmentation outputs but using diverse estimation strategies. All top three teams relied on 3D models, using CNN, U-Net, and Transformer architectures. Despite this methodological diversity, most submissions failed to outperform a simple GA-based linear regression baseline, highlighting that automated fetal brain biometry is still an open problem. 

Several limitations can be observed across these studies. First, many approaches are restricted to a limited set of biometric measurements, typically focusing on only a few clinically relevant measurements  \cite{avisdris2021automatic,she2023automatic,gong2025fetal}. Second, a number of methods are evaluated exclusively on HC \cite{she2023automatic,luis2026towards}, which limits conclusions regarding robustness in pathological anatomy. Finally, although more comprehensive morphometry frameworks exist \cite{masterl2024enhancing}, these rely on earlier-generation pipelines and do not reflect recent advances in deep learning-based segmentation and reconstruction. Overall, systematic evaluation of modern architectures across both HC and PC, as well as comprehensive multi-metric biometric extraction, remains insufficiently explored.

\begin{figure}[!t]
    \centering
    \includegraphics[width=1.0\linewidth]{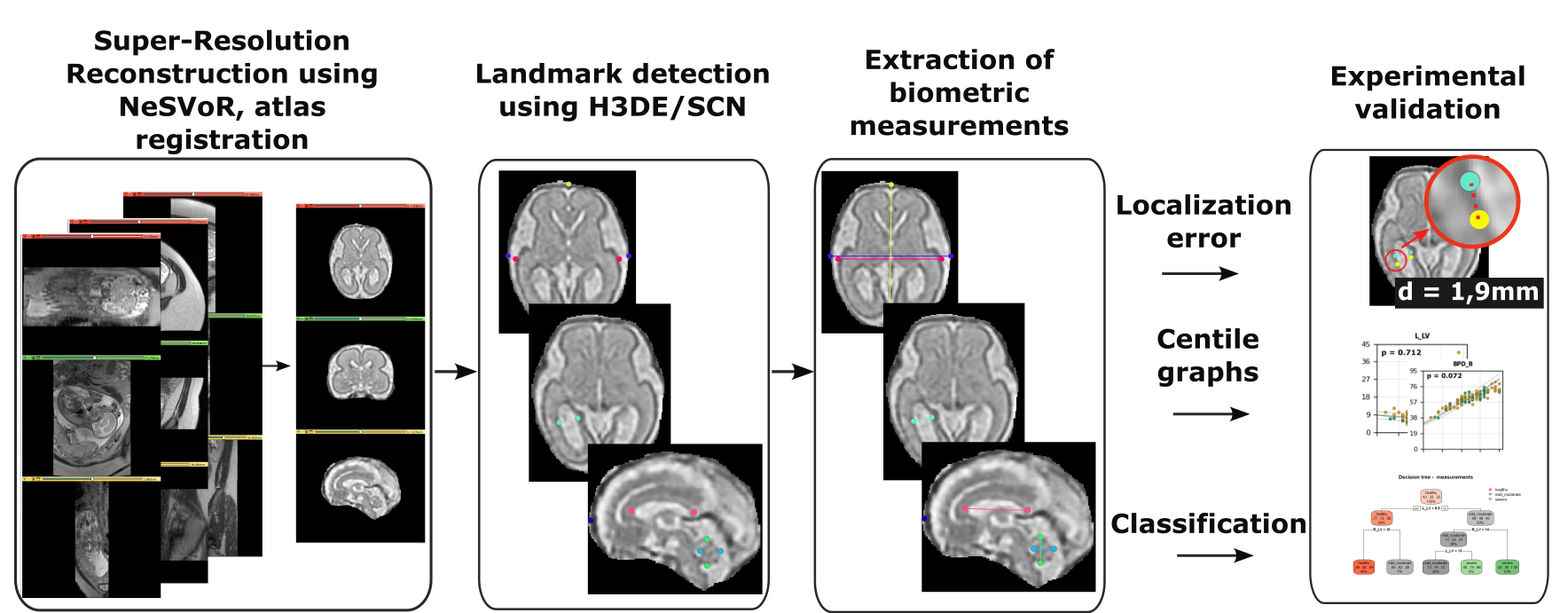}
    \caption{The proposed fetal brain MRI biometry pipeline.}
    \label{pipeline}
\end{figure}

In this paper, we assemble and evaluate an automated biometric analysis pipeline for fetal brain MRI  with the following contributions: (1) the use of anatomical landmark localization using two models, H3DE-Net \cite{huang2025h3de} and Spatial Configuration Net (SCN) \cite{payer2019integrating}, from which we derive 11 biometric measurements covering supratentorial, ventricular, cerebellar, and midline structures (from 22 landmarks); (2) a statistical comparison of the two models' localization accuracy across landmarks and diagnostic groups on a heterogeneous cohort comprising both HC and a range of pathologies, rather than HC only; (3) an assessment of agreement between the derived measurements and normative growth trajectories using calibrated centile charts, quantifying deviations of PC from the reference curves; (4) comparison of the biometric results to a GA-based linear regression model and (5) a classification analysis distinguishing severity grades of ventriculomegaly (VM) from the derived measurements. The study design is shown in Fig.~\ref{pipeline}.

\section{Materials and Methods}
\subsection{Dataset description}
The dataset consisted of 122 fetal MRI acquisitions with gestational age (GA) from 18 to 40 weeks (mean 31±4 weeks). GA was also comparable across both HC and PC groups (32 ± 3 vs. 31 ± 4 weeks). Inclusion criteria were: (1) only singletone pregnancies, (2) absence of excess fetal motion or artifacts, (3) image stacks captured whole fetal brain and (4) no cases with complete agenesis of structures such as corpus callosum or cavum septum pellucidum. The cohort comprised fetuses with normally developing brain (53) and those diagnosed various brain anomalies: VM (63), cerebellar pathologies (8), corpus callosum pathologies (8), genetic mutations (4), and hypogyration (3). Some cases have multiple associated pathologies. 

Imaging was performed at the University Medical Center Ljubljana using a 1.5 T clinical MRI scanner (Siemens Aera, Siemens Healthineers, Erlangen, Germany) with a T2-weighted Half-Fourier Acquisition Single-shot Turbo spin-Echo (HASTE) sequence. Each examination consisted of at least three orthogonal stacks covering the full fetal brain. In cases of visible motion artifacts during screening, the affected stack was reacquired to ensure adequate image quality. Acquisition parameters provided high in-plane resolution (0.625 × 0.625 mm) with 3 mm slice thickness, yielding 512 × 320 matrices and 31–35 slices per stack. All datasets were anonymized prior to analysis.

The study protocol received approval from the Institutional Review Board (IRB; approval no.: 0120-56/2022/3). The IRB waived collection of informed consents for this retrospective, secondary data analysis study.  

For each fetus, 2D MRI stacks were reconstructed into a 3D volume using the NeSVoR algorithm~\cite{xu2023nesvor}, resulting in 0.5 mm isotropic resolution. This reconstruction method was selected based on findings in our previous work - Masterl et al. \cite{masterl2025robustness}, where NeSVoR achieved the highest rate of successful reconstructions. To ensure a consistent anatomical reference across the cohort, all reconstructed volumes were then aligned to an age-matched brain atlas using rigid registration~\cite{uus2023multi}.

\subsection{Landmark localization}
\begin{figure}
    \centering
    \includegraphics[width=0.9\linewidth]{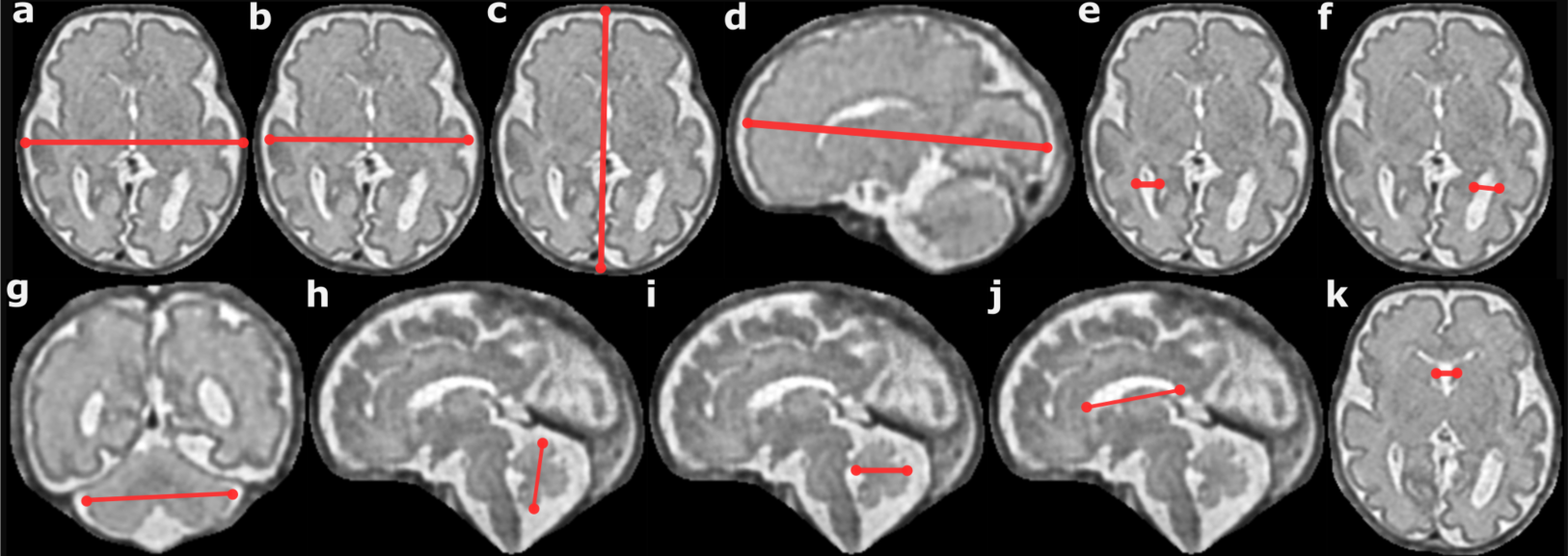}
    \caption{Biometric measurements: a--skull biparietal diameter (BPD\_S), b--brain biparietal diameter (BPD\_B), c--skull occipito-frontal diameter (FOD\_S), d--brain occipito-frontal diameter (FOD\_B), e--right lateral ventricle width (R\_LV), f--left lateral ventricle width (L\_LV), g--transcerebral diameter (TCD), h--vermis height (VH), i--vermis diameter (VD), j--cavum septum pellucidum width (CSP), k--corpus callosum length (CCL).}
    \label{measurements}
\end{figure}

On reconstructed 3D fetal brain volumes we aimed to predict 22 landmarks in order to derive 11 clinically relevant biometric measurements. The set of relevant measurements was determined  by an experienced radiologist with more than 15 years of experience. Both the landmarks and the resulting measurements are shown in Fig.~\ref{measurements}. Manual landmark annotation was performed using the 3D Slicer Markups tool by an image analysis practitioner with four years of experience in fetal brain MRI interpretation, who had been trained by the same radiologist responsible for defining the measurement protocol.

We trained two landmark detection models, SCN, which served as the localization model in our previous work \cite{masterl2024enhancing}, and the more recent H3DE-Net, to assess whether a newer architecture improves upon our established approach, with evenly distributed HC and PC across folds.  Dataset was distributed into five similarly sized folds, where all folds but one were used for training. In the training part of the dataset, 20\% of the cases were used for validation during training and to manage early stopping. H3DE-Net was trained with the Adam optimizer and HNP heatmap loss function (learning rate 0.001, weight decay 0.0005, betas 0.9/0.98). SCN was trained with AdamW and Gaussian Heatmap Loss (learning rate 1e-4, weight decay 1e-3 for network parameters). For both models, the optimizer, learning rate, and weight decay were selected on the validation split to yield the best localization performance for each architecture individually, and were then fixed across all folds. Both used an exponential learning rate decay (gamma=0.98) applied until the learning rate reached 3\% of its initial value, and early stopping based on validation loss. Due to smaller dataset we also included data augmentation in training part. Augmentation consisted of random affine transformations (rotation and scaling). Rotation angles were sampled uniformly from $\pm$0.1 rad ($\approx \pm$5.7$^\circ$) to reflect the limited orientation variability expected in atlas-registered data. Scaling factors were sampled uniformly from 0.9 to 1.1 ($\pm$10\%) to approximate the volumetric variation associated with fetal brain growth.

\section{Experimental validation}
All results reported in the following analyses were obtained on the held-out test folds, so that every case was evaluated by a model that had not seen it during training. 
\subsection{Setup}
\subsubsection{Registration consistency across diagnostic groups:}
Registration differences between HC and PC cases could lead to different landmark localization results. To evaluate whether such differences were present, we calculated two image similarity metrics, the peak signal-to-noise ratio (PSNR) and the structural similarity index measure (SSIM). We then compared these metrics between the two diagnostic groups using Mann-Whitney U test.

\subsubsection{Landmarks localization accuracy:}
To characterize the differences in localization accuracy between H3DE-Net and SCN across landmarks and diagnostic groups, we first computed the Euclidean distance between predicted and reference landmark coordinates. Furthermore, to account for repeated measurements within the same case, a linear mixed-effects model was fitted to the localization error using R (v4.4.2). The model was estimated using the \texttt{lme4} package (v2.0.1), with auxiliary mixed-model tools from  \texttt{lmerTest} (v3.2.1). The model included \texttt{Model}, \texttt{Landmark}, and \texttt{Diagnosis} as fixed effects, along with \texttt{Model × Landmark and Model × Diagnosis} interactions, and a random intercept for each case. Significance of fixed effects was assessed using a Type III ANOVA implemented via the \texttt{car} R package (v3.1.5). Pairwise comparisons between models were obtained from estimated marginal means computed using the \texttt{emmeans} R package (v2.0.3) with Tukey adjustment for multiple comparisons, and results are reported as error ratios with 95\% confidence intervals. Model assumptions were evaluated using the \texttt{DHARMa} R package (v0.4.7).

\subsubsection{Agreement with normative values:}
To assess whether the empirical measurements aligned with the predicted growth trajectories and to identify deviations of PC data points from normative reference curves, we utilized adapted centile charts derived from Luis et al. \cite{luis2026towards}. The curves were vertically shifted to calibrate for systematic offsets between our cohort and the reference population. The assumption of a uniform distribution of measurements for HC was verified using the Kolmogorov-Smirnov (KS) test. To control the family-wise error rate, $p$-values were adjusted for multiple comparisons using Hommel's method. Furthermore, we quantified extreme deviations by evaluating the proportion of observations falling outside the 5th--95th centile interval.

\subsubsection{Comparison with GA-based linear regression:}
To compare the results with findings of FeTA challenge we also developed GA-based linear regression model and calculated mean average percent error (MAPE) for both the results of biometry models and regression model. When training regression model we performed 5-fold CV, where data was split with respect to diagnosis. The CV was repeated 100 times. We repeated the experiment with only HC cases, to determine whether the diagnosis impacts the results.

\subsubsection{Diagnostic classification:}
To accommodate the data volume within each diagnostic group, the analysis was restricted to HC and PC with prevalent VM. The diagnosis of VM and its categorical definition were rigorously established based on LV width. To evaluate the capacity of model-derived measurements to accurately infer diagnostic categories and to identify optimal diagnostic thresholds, a decision tree classifier was implemented. To ensure valid paired comparisons across methods, identical cases, class labels (HC = 52, mild/moderate = 39, severe = 31; mild and moderate cases were pooled due to limited sample sizes), and cross-validation fold assignments were maintained. Model performance was robustly evaluated using ten-fold cross-validation, with uncertainty estimates quantified via bootstrap resampling over 2000 iterations. To prevent overfitting, the maximum tree depth was constrained to 2 based on visual inspection of the initial results.

\subsection{Results}
\subsubsection{Registration consistency across diagnostic groups:}
Registration consistency was comparable between the two diagnostic groups for both metrics. SSIM did not differ between groups (HC: 0.740 ± 0.076; PC: 0.738 ± 0.097; Mann–Whitney U, p = 0.99; Cohen's d = 0.03), nor did PSNR (HC: 11.20 ± 0.57 dB; PC: 11.11 ± 0.81 dB; Mann–Whitney U, p = 0.71; Cohen's d = 0.12). In both cases the p-values were well above 0.05 and the effect sizes were negligible. The median values were likewise close between groups.

\subsubsection{Landmark localization errors:}

Table \ref{mre} summarizes the landmark localization accuracy of both models separated by HC and PC. The H3DE-Net achieved the lowest mean localization error for VH and VD and the highest for BPD\_B, while the error was generally higher in the PC group than in the HC group. The SCN likewise achieved its lowest localization error for VH and its highest for BPD\_B; however, no consistent differences in localization error were observed between the HC and PC groups. The standard deviation was also generally higher in the PC group for both models.

Mixed-effects model analysis showed that the H3DE-Net achieved significantly lower localization errors than the SCN across landmarks, with error ratios (SCN/H3DE-Net) ranging from about 1.5 to 3.9 and all p-values below 0.001 after correction for multiple comparisons. The Type III ANOVA revealed a significant Model × Landmark interaction (HC: F = 4.96; PC: F = 5.86; both p < 0.001), indicating that this advantage varied in magnitude across landmarks rather than being constant. The H3DE-Net also exhibited a statistically significant increase in localization error in the PC group compared with the HC group, whereas no significant difference between the two groups was observed for the SCN.

\begin{table}[htbp]
\centering
\scriptsize
\begin{tabular}{|c|c|c||c|c|}
\hline
{\textbf{H3DE-Net}}& \multicolumn{2}{c||}{Diagnosis HC} & \multicolumn{2}{c|}{Diagnosis PC} \\
\hline
Measurements (mm) & ps accuracy & pe accuracy & ps accuracy & pe accuracy \\
\hline
BPD\_S & 1.59 ± 0.67 & 1.90 ± 0.61 & 2.38 ± 3.92 & 2.53 ± 2.34 \\
BPD\_B & 2.46 ± 2.39 & 1.88 ± 1.07 & 2.28 ± 1.79 & 2.34 ± 3.11 \\
FOD\_S & 1.62 ± 0.83 & 1.72 ± 1.19 & 1.91 ± 1.02 & 1.98 ± 2.28 \\
FOD\_B & 1.77 ± 1.22 & 1.26 ± 1.24 & 2.20 ± 2.37 & 1.71 ± 2.00 \\
R\_LV & 1.13 ± 0.52 & 1.40 ± 0.91 & 2.08 ± 1.99 & 2.22 ± 1.90 \\
L\_LV & 1.17 ± 0.60 & 1.25 ± 0.71 & 2.73 ± 3.50 & 2.11 ± 2.56 \\
TCD & 1.17 ± 1.67 & 1.21 ± 2.06 & 1.39 ± 2.09 & 1.42 ± 1.83 \\
VH & 0.96 ± 0.47 & 0.80 ± 0.41 & 1.45 ± 1.86 & 1.35 ± 2.06 \\
VD & 0.87 ± 0.66 & 1.10 ± 0.54 & 1.21 ± 1.62 & 2.14 ± 3.36 \\
CCL & 1.18 ± 0.63 & 1.13 ± 0.80 & 1.80 ± 3.50 & 1.58 ± 1.48 \\
CSP & 1.14 ± 0.61 & 1.19 ± 0.64 & 1.51 ± 1.14 & 1.51 ± 1.11 \\
\hline
Average & \multicolumn{2}{c||}{1.36 ± 1.12} & \multicolumn{2}{c|}{1.90 ± 2.38} \\
\hline\hline

{\textbf{SCN}} & \multicolumn{2}{c||}{Diagnosis HC} & \multicolumn{2}{c|}{Diagnosis PC} \\
\hline
Measurements (mm) & ps accuracy & pe accuracy & ps accuracy & pe accuracy \\
\hline
BPD\_S & 3.20 ± 1.86 & 3.66 ± 2.26 & 3.98 ± 5.48 & 3.86 ± 3.05 \\
BPD\_B & 4.50 ± 3.63 & 4.70 ± 2.13 & 4.68 ± 6.30 & 4.95 ± 4.67 \\
FOD\_S & 4.91 ± 3.91 & 2.95 ± 2.12 & 4.31 ± 3.09 & 3.89 ± 3.88 \\
FOD\_B & 3.08 ± 1.90 & 5.18 ± 3.58 & 3.44 ± 2.80 & 4.62 ± 3.73 \\
R\_LV & 3.20 ± 1.77 & 3.31 ± 2.16 & 4.28 ± 5.15 & 4.75 ± 4.22 \\
L\_LV & 3.07 ± 1.75 & 3.15 ± 1.83 & 5.32 ± 7.20 & 4.36 ± 7.14 \\
TCD & 3.60 ± 2.46 & 3.91 ± 3.34 & 3.85 ± 5.14 & 5.86 ± 7.60 \\
VH & 2.92 ± 1.78 & 3.25 ± 2.25 & 2.93 ± 2.84 & 3.48 ± 2.98 \\
VD & 3.04 ± 1.90 & 3.26 ± 1.85 & 3.24 ± 4.79 & 4.59 ± 6.70 \\
CCL & 3.55 ± 2.50 & 2.88 ± 1.74 & 3.75 ± 4.82 & 3.54 ± 3.92 \\
CSP & 3.80 ± 3.60 & 3.69 ± 2.37 & 3.69 ± 4.88 & 3.42 ± 2.48 \\
\hline
Average & \multicolumn{2}{c||}{3.58 ± 2.56} & \multicolumn{2}{c|}{4.13 ± 4.94} \\
\hline
\end{tabular}
\caption{Mean and standard deviation of the Euclidean distances measuring startpoint (ps) and endpoint (pe) localization error for H3DE-Net and SCN, reported separately for the HC and PC groups.}
\label{mre}
\end{table}

\subsubsection{Comparison with GA-based linear regression:}

Results from Table \ref{mape}  shows that H3DE-Net outperforms the GA baseline in 7 of 11 measurements, failing to surpass it only for the smaller structures VH, VD, CCL and CSP. SCN outperforms the baseline only for the LV. On all other structures the baseline is more accurate than both trained models, with the largest gap on the smaller measurements. The advantage of the trained models is concentrated at the LV, where the baseline error is high.

The split by diagnosis shows that the baseline's LV error is driven by the PC cases. For L\_LV and R\_LV the baseline MAPE is markedly lower in HC than in PC, so H3DE-Net's advantage over the baseline narrows in HC and widens in PC. For the remaining structures the relative ordering is stable across subsets: the baseline remains superior for VH, VD, CCL and CSP, and H3DE-Net remains superior for the biparietal, occipito-frontal and transcerebellar measurements regardless of diagnosis.

\begin{table}[]
\centering
\scriptsize
\caption{MAPE results for the GA based regressiom (GA) and both trained models across all cases, HC only, and PC only. Bold marks measurements where the trained model outperformed the baseline.}
\label{mape}
\setlength{\tabcolsep}{4pt}
\begin{tabular}{|l|c|c|c|c|c|c|c|c|c|}
\hline
& \multicolumn{3}{c|}{All cases} & \multicolumn{3}{c|}{HC only} & \multicolumn{3}{c|}{PC only} \\
\cline{2-10}
Measurement & GA & SCN & H3DE-Net & GA & SCN & H3DE & GA& SCN & H3DE-Net \\
\hline
BPD\_S & 3,62 & 7,58 & \textbf{1,88} & 4,11 & 7,00 & \textbf{1,34} & 3,42 & 8,01 & \textbf{2,28} \\ \hline
BPD\_B & 4,26 & 11,51 & \textbf{2,50} & 3,67 & 9,75 & \textbf{2,21} & 4,23 & 12,84 & \textbf{2,72} \\ \hline
FOD\_S & 3,20 & 6,65 & \textbf{1,61} & 2,93 & 6,50 & \textbf{1,43} & 3,39 & 6,75 & \textbf{1,74} \\ \hline
FOD\_B & 2,73 & 8,68 & \textbf{2,18} & 2,21 & 8,76 & \textbf{2,07} & 3,15 & 8,62 & \textbf{2,26} \\ \hline
L\_LV & 49,17 & \textbf{25,45} & \textbf{20,07} & 27,85 & \textbf{24,41} & \textbf{21,60} & 36,22 & \textbf{26,22} & \textbf{18,91} \\ \hline
R\_LV & 50,70 & \textbf{29,90} & \textbf{23,88} & 34,48 & \textbf{31,37} & \textbf{25,71} & 36,85 & \textbf{28,79} & \textbf{22,51} \\ \hline
TCD & 6,66 & 14,51 & \textbf{5,67} & 4,11 & 12,50 & \textbf{3,75} & 8,43 & 16,02 & \textbf{7,11} \\ \hline
VH & 7,34 & 15,45 & 10,09 & 6,14 & 14,36 & 8,43 & 7,96 & 16,27 & 11,33 \\ \hline
VD & 13,04 & 20,53 & 20,14 & 10,50 & 17,28 & 17,90 & 15,14 & 22,97 & 21,83 \\ \hline
CCL & 8,95 & 10,43 & 12,67 & 7,04 & 8,30 & 11,94 & 10,57 & 12,03 & 13,22 \\ \hline
CSP & 30,43 & 37,52 & 30,81 & 22,98 & 29,91 & 23,35 & 35,41 & 43,23 & 36,41 \\
\hline
\end{tabular}
\end{table}

\subsubsection{Agreement with normative values:}
\begin{figure}
    \centering
    \includegraphics[width=0.89\linewidth]{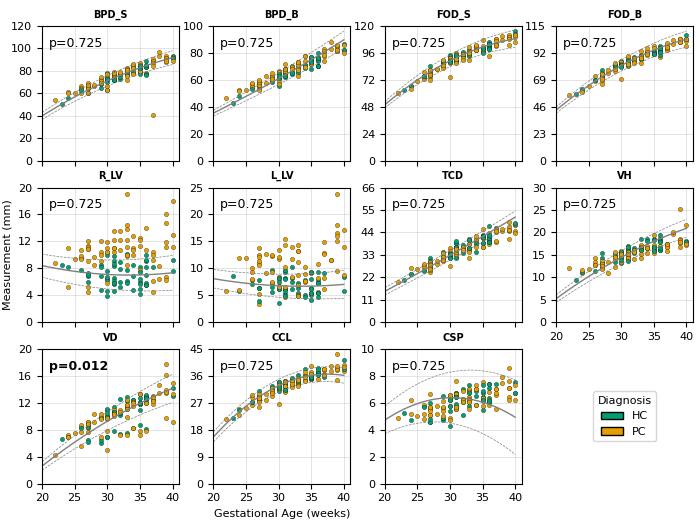}
    \caption{Anatomical measurements with GA, 5th, 50th and 95th percentile curves. Cases are divided by the diagnosis. The p-values correspond to KS uniformity test performed on HC .}
    \label{centiles}
\end{figure}

Fig. \ref{centiles} shows biometric measurements for all 122 cases with respect to GA and colored labels dividing HC and PC for results of H3DE-Net.
The KS test on the results obtained with the H3DE-Net showed the distribution of all of the measurements except the VD was uniform. The proportion of cases falling outside the 5th–95th centile range differed clearly between the two groups. In the HC group the proportions were modest and, for most measurements, reasonably balanced between the two tails, consistent with the near-uniform distribution. The exceptions were VD (16.7\% above, 20.4\% below), TCD (14.8\% above, 16.7\% below) and VH (16.7\% above, 13.0\% below), where a larger share of cases lied outside the expected range. In the PC group the pattern was strongly asymmetric and shifted towards the upper tail: 55.6\% of L\_LV and 45.8\% of R\_LV cases exceed the 95th centile, consistent with the predominance of VM among PC. BPD\_B and BPD\_S were also shifted upward (20.8\% and 13.9\% above the 95th centile). TCD showed the opposite tendency, with 22.2\% of PC cases below the 5th centile, while VD was split across both tails (12.5\% above, 16.7\% below). FOD\_S, FOD\_B, CCL and CSP remained close to the expected range in both groups.

For the results of SCN, the KS test rejected uniformity of the HC centile distribution for BPD\_B (p = 0.006), TCD (p = 0.01) and CSP (p = 0.03), the remaining measurements were consistent with a uniform distribution. The proportion of HC outside the 5th–95th centile range was, however, elevated for a larger set of measurements than the KS test alone indicates. BPD\_S (18.5\% above), BPD\_B (25.9\% above, 13.0\% below), L\_LV (35.2\% above), R\_LV (22.2\% above), TCD (27.8\% above, 22.2\% below) and VH (14.8\% above, 25.9\% below) all show a substantial share of HC outside the expected range, exceeding the \~10\% expected under correct calibration. FOD\_S, FOD\_B, VD, CCL and CSP remained close to the expected range in the HC group. In the PC group the upper-tail shift at the ventricles was present (38.9\% of L\_LV and 30.6\% of R\_LV above the 95th centile). TCD again showed a downward tendency (16.7\% of PC  below the 5th centile).

\subsubsection{Classification models}
For both models the decision tree selected only ventricular width. Using measurements obtained with the H3DE-Net, the decision tree split on L\_LV at 9.13 mm at the root, separating HC from VM, followed by R\_LV at 10.61 mm (HC vs. mild/moderate) on the left branch and L\_LV at 11.93 mm (mild/moderate vs. severe) on the right. With the use of SCN based measurements, the decision tree had the same structure with lower thresholds (L\_LV 7.955 mm at the root; R\_LV 9.43 mm and L\_LV 13.15 mm at the second level). In both cases the thresholds separating HC from VM (7.955–9.13 mm) and mild/moderate from severe (11.93–13.15 mm) fall near the clinical 10 mm and 15 mm cut-offs used to define and grade VM. 
Across all three diagnostic groups, the H3DE-Net achieved higher AUC values than the SCN, as shown in Table \ref{classification}. For the HC group, the confidence intervals of the two methods do not overlap, indicating a clear performance difference. In contrast, the confidence intervals for the mild/moderate and severe groups partially overlap, suggesting less pronounced differences between the models. Both methods achieved the highest discriminative performance for the HC group, whereas the mild/moderate group proved to be the most challenging to classify.
The DeLong test revealed a statistically significant difference in AUC between H3DE-Net and SCN only for the HC group ($p = 0.04$). No statistically significant differences were observed for the mild/moderate ($p = 0.21$) or severe ($p = 0.06$) groups.
A paired case-level comparison using McNemar's test also demonstrated a statistically significant difference in overall classification performance ($p = 0.0034$). H3DE-Net correctly classified 31 cases that were misclassified by SCN, whereas SCN correctly classified 11 cases that were misclassified by H3DE-Net. Both methods misclassified the same 13 cases.

\makeatletter
\newcommand{\thickhline}{\noalign{\ifnum0=`}\fi\hrule \@height 1pt \futurelet\reserved@a\@xhline}
\makeatother
 
\begin{table}[t]
\centering
\scriptsize
\caption{Classification performance for SCN and H3DE-Net}
\label{classification}
\begin{tabular}{|l|l|c|c|c|}
\hline
Method & Class & AUC & Sensitivity & Specificity \\
\hline
\multirow{3}{*}{H3DE}
 & HC            & 0.93 (0.87--0.97)& 0.92 (0.84--0.98)& 0.95 (0.90--1.00)\\ \cline{2-5}
 & Mild/moderate & 0.73 (0.63--0.82)& 0.84 (0.71--0.94)& 0.78 (0.69--0.86)\\ \cline{2-5}
 & Severe        & 0.86 (0.80--0.91)& 0.54 (0.38--0.71)& 0.96 (0.92--1.00)\\
\thickhline
 \multirow{3}{*}{SCN}
 & HC            & 0.84 (0.76--0.92)& 0.78 (0.67--0.88)& 0.94 (0.88--0.98)\\ \cline{2-5}
 & Mild/moderate & 0.65 (0.55--0.74)& 0.69 (0.53--0.82)& 0.62 (0.54--0.72)\\ \cline{2-5}
 & Severe        & 0.75 (0.66--0.84)& 0.32 (0.16--0.48)& 0.90 (0.84--0.95)\\
\hline
\end{tabular}
\end{table}

\section{Discussion}

This study evaluated an automated biometric analysis pipeline for fetal brain MRI on a heterogeneous cohort of HC and PC cases, comparing two landmark localization models (H3DE-Net and SCN) in terms of localization accuracy, agreement with normative values, comparisson to GA-based regression, and classification of VM severity.

Localization was generally worse for PC than HC. For H3DE-Net this difference was significant, whereas for SCN it was not; the most plausible explanation is that SCN's higher baseline error and variance mask the effect rather than indicating robustness to the presence of pathology. This has a direct methodological implication: models trained and evaluated only on HC, as in prior studies \cite{she2023automatic,luis2026towards}, cannot be assumed to transfer to pathological anatomy, and their performance in that setting has to be established explicitly rather than inferred from HC results.

The agreement with normative values was consistent with the localization results. For the H3DE-Net the HC measurements followed the expected centile distribution, with VD as the only exception, whereas for the SCN several measurements deviated and a larger share of HC fell outside the reference range, indicating poorer calibration on healthy cases. In the PC group both models reproduced the expected upward shift at the lateral ventricles, consistent with the predominance of VM. We can also observe that for CSP, both HC and PC cases deviate significantly from expected distribution. We hypothesize that this reflects CSP being the smallest of all eleven measurements: for a small structure, measurement errors and discrepancies arising from differing annotation protocols translate into proportionally larger deviations, which become visually apparent even when the points remain within the reference bounds.

Our dataset includes a range of pathologies, most notably cases with pronounced VM. Here the advantage of the landmark localization models becomes apparent, as a GA-only predictor cannot capture pathological deviation from normative growth. A natural extension would be to reproduce this analysis on the publicly available Kispi dataset, applying our measurement set and diagnostic stratification, which were not part of the original challenge evaluation. 

Separating HC from VM was the most clinically relevant distinction and the one where the models differed most, with the measurements obtained using the H3DE-Net performing best. Finer stratification was more limited: the mild/moderate group was the hardest to classify and mild and moderate cases had to be pooled because of small sample sizes. Reliable classification into severity grades will therefore require collecting more cases per diagnostic group.

Several limitations remain. While our intended application is the diagnosis of multiple pathologies, the classification analysis had to be restricted to HC and VM, as the other pathologies were too few in numbers. Future work also includes adjusting the models to account for cases with agenesis of structures such as the corpus callosum or cavum septum pellucidum, which were excluded here. Recent years have also seen the development of other landmark localization algorithms, such as nnLandmark by Ertl et al. ~\cite{ertl2026nnlandmark}. Evaluating such methods within our pipeline would allow a broader comparison beyond the two models considered here. 

\section{Acknowledgements}
This work was supported by the Slovenian Research Agency (Core Research Grant No. P2-0232 and Research Grants Nos. J2-2500 and J2-3059).

\bibliographystyle{splncs04}
\bibliography{main}

\end{document}